\pdfoutput=1
\documentclass[11pt]{article}

\usepackage{EACL2023}

\usepackage{times}
\usepackage{latexsym}
\usepackage{graphicx} 

\usepackage[T1]{fontenc}
\usepackage[utf8]{inputenc}

\usepackage{microtype}

\usepackage{inconsolata}

\title{Uncovering the Handwritten Text in the Margins: End-to-end Handwritten Text Detection and Recognition}

\author{Liang Cheng\textsuperscript{1,*}, Jonas Frankemölle\textsuperscript{2,*}, Adam Axelsson\textsuperscript{1,*} and Ekta Vats\textsuperscript{1,$\dagger$}\\
  \textsuperscript{1}Department of Information Technology, Uppsala University, Sweden \\
  \textsuperscript{2}Department of Archives, Libraries and Museums, Uppsala University, Sweden \\
  \textsuperscript{$\dagger$}\texttt{ekta.vats@it.uu.se}}

\begin{document}
\maketitle

\def\thefootnote{*}\footnotetext{Equal contribution}\def\thefootnote{\arabic{footnote}}

\begin{abstract}
The pressing need for digitization of historical documents has led to a strong interest in designing computerised image processing methods for automatic handwritten text recognition. However, not much attention has been paid on studying the handwritten text written in the margins, i.e. marginalia, that also forms an important source of information. Nevertheless, training an accurate and robust recognition system for marginalia calls for data-efficient approaches due to the unavailability of sufficient amounts of annotated multi-writer texts. Therefore, this work presents an end-to-end framework for automatic detection and recognition of handwritten marginalia, and leverages data augmentation and transfer learning to overcome training data scarcity. The detection phase involves investigation of R-CNN and Faster R-CNN networks. The recognition phase includes an attention-based sequence-to-sequence model, with ResNet feature extraction, bidirectional LSTM-based sequence modeling, and attention-based prediction of marginalia. The effectiveness of the proposed framework has been empirically evaluated on the data from early book collections found in the Uppsala University Library in Sweden. Source code and pre-trained models are available at Github\footnote{\url{https://github.com/adamaxelsson/Project-Marginalia}}.

%To the best of author's knowledge, this is the first work in the community that presents computational methods for reading marginalia.

\end{abstract}

\section{Introduction}
Libraries and archives across the globe are in possession of rich cultural heritage collections to be digitized for preservation and preventing degradation over time. For example, the Uppsala University Library in Sweden maintains several early book collections, dating back to the 1400s. An example of such a collection is the Walleriana book collection, encompassing medicine and science \citep{uubWaller}. These collections are an important source of evidence for the European history and are valuable for researchers. Much of the content from these collections is well documented and is available online. 

However, many books and documents, in addition to printed text, contain handwritten marginalia i.e. text written in the margins. This marginalia are also an important source of information for researchers, but is not as voluminous as the printed text. The presence of marginalia is sometimes mentioned, but its content is not. This is also due to poor readability of handwritten marginalia texts and challenges such as high variability in different writing styles, languages and scripts. Therefore, it is of great value to develop computational methods for digitization of the handwritten marginalia to make it as available as the printed text of these collections.

%Given an input scanned document image, the algorithm involves automatic detection of text in the margins, line segmentation, word segmentation, and lastly, recognition at word-level. 

\begin{figure*}[!htbp]
\centering
\includegraphics[width=6in]{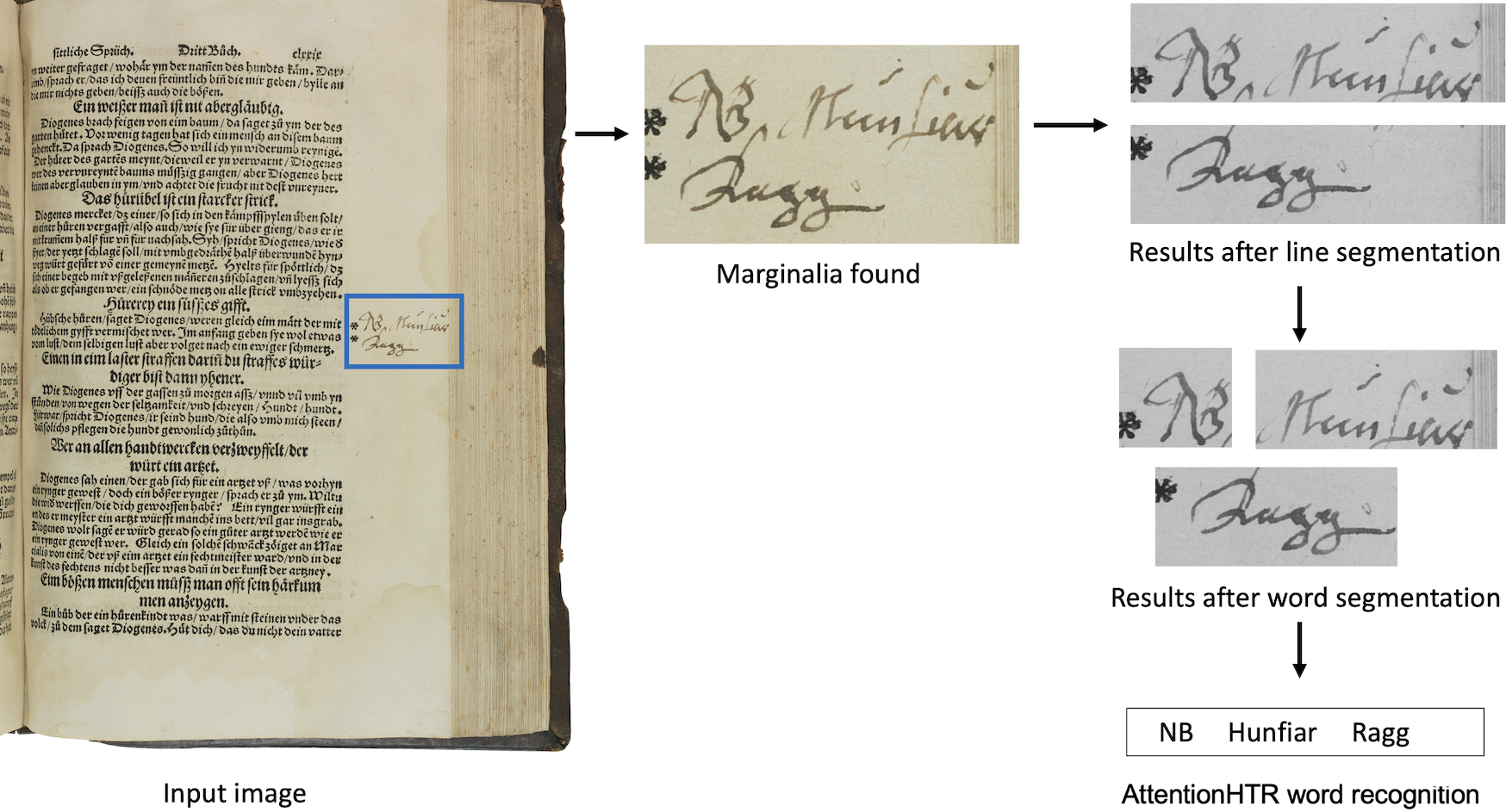}
\caption{Overall pipeline of the end-to-end marginalia detection and recognition framework. The recognition result can be further improved using a suitable language model.}
\label{fig:pipeline}
\end{figure*}

To do so, this work presents an end-to-end deep learning based approach for handwritten marginalia detection and recognition. Two different deep learning architectures: Region-based Convolutional Neural Network (R-CNN) and (Faster R-CNN) are studied for marginalia detection. The aim is for the networks to predict the coordinates of handwritten marginalia based on an input document image. To digitize the contents of the marginalia, there needs to be a way to automatically read it. To achieve this, an algorithm for segmenting handwritten text to individual words is also presented. Finally, when the marginalia have been identified and segmented, each word is fed in to an attention-based encoder-decoder network for handwritten text recognition (HTR), i.e. AttentionHTR introduced in our previous work in \citep{kass2022attentionhtr}. The encoder block constitutes ResNet feature extraction and bidirectional LSTM-based sequence modeling stages, and the prediction stage consists of a decoder and a content-based attention mechanism. 

To train the marginalia detector network, a sample dataset of 513 scanned pages from Uppsala University Library's book collections is used. The dataset was labelled by an expert as part of this work using an open-source tool LabelMe \citep{russell2008labelme}, and the data thus obtained contains labeled coordinates for the marginalia, which are used as the targets for the network training. The overall pipeline of the end-to-end- marginalia detection and recognition framework is presented in Figure \ref{fig:pipeline}. 

To the best of authors' knowledge, this is the first attempt at extracting the old historical handwritten text in the margins using an end-to-end detection and recognition pipeline. The research is reproducible, with user-friendly and modular designed code, and provide scope for future research and exploration of marginalias. 
%In order to evaluate the performance of thr two networks, \textit{Intersection over Union}, or \textit{IoU} is used. 

\section{Related Work}
%Attempting to automatically read historical handwritten documents is not a new idea, several similar projects have already been done. One example is the report \textit{Boosting Handwriting Text Recognition in Small Databases with Transfer Learning}\citep{aradillas2020improving}. In this report the authors discuss the ... 

Since the advent of deep neural networks, the HTR research has witnessed significant advancement in method design and development, with popular approaches such as Transformers based architectures TrOCR \citep{li2023trocr} and attention-based sequence to sequence models \citep{kass2022attentionhtr, bluche2017scan, kang2019convolve}. Our previous work \citep{kass2022attentionhtr} introduced an end-to-end HTR system based on attention encoder-decoder networks, where the attention-based architecture is simple, modular, and reproducible, allowing more data to be added in the pipeline. 

There have been other attempts at automatically reading historical handwritten documents \citep{nockels2022understanding}. For example, \citep{aradillas2020improving} discusses the challenge of dealing with different styles of handwriting. Since two documents can be from completely different centuries and countries, there is bound to be a lot of variability in the handwriting. The solution that is presented by the authors is to use transfer learning. First, a network is trained on a large set containing handwritten text of modern English. This base network can then be tweaked, depending on what kind of text will be processed. For example, if the network is to be used for a certain collection, a subset of this collection can be used to further train the network. Doing this can make the network significantly better at making predictions on that collection. The benefit of this technique is that since some collections can be very small, it can be difficult to train a network solely on that collection without running the risk of over-fitting. Transfer learning handles this problem by first creating a good general model, that is then tweaked and specialized to a certain dataset. 

Another related work \citep{bluche2016joint} presents a joint line segmentation and transcription approach for end-to-end handwritten paragraph recognition. To do so, handwritten text is automatically segmented into individual lines by using Multi-Dimensional Long Short-Term Memory Recurrent Neural Networks, or MDLSTM-RNN for short. 

However, the problem of reading the text in the margins is relatively under-explored \citep{goodwin2021locating}. The work \citep{bold2017marginalia} highlighted the importance of marginal annotations for the community and how it benefits the readers and historians. For example, for historians working with texts, marginalia can provide insights into earlier readers and their perspectives. There have been some attempts at studying the marginalia notes and drafts by Moby-Dick author Herman Melville in \citep{ohge2018axis, lambie2022visualizing, hitchcock2023gender}. These works focus on the visualization of marginalia, gender research and knowledge exploration for instance. On the other hand, our proposed work focuses on developing advanced computational methods for automatic detection and recognition of marginalia, leveraging modern deep learning models and our reproducible research.

%Reviewer 3: I wonder about its similarities or overlaps with recent neural net-based page segmentation and OCR tools like Eynollah. It would be good to know how, despite its originality, this pipeline fits into a larger ecosystem of approaches for text recognition on historical documents. The authors offer a few examples of why reading marginalia is valuable to book historians, and I think it's possible to press this point even further. Mia Goodwin's "Locating Digitised Marginalia" and the whole *Marginal Notes* collection from 2021 would be a good touchstone for! why marginalia is important and how difficult it currently is to find and examine marginal text digitally.

%is \textit{Joint Line Segmentation and Transcription for End-to-End Handwritten Paragraph Recognition}\citep{bluche2016joint}. This report is about automatically segmenting handwritten text to individual lines by using a neural network called multi-dimensional long short-term memory recurrent neural networks, or MDLSTM-RNN for short. Similar to our work, the text is segmented to lines. These lines are then fed to an HTR model. We will add another step in our segmentation, which is segmenting each line to individual words before feeding it to the HTR model.

\begin{figure*}[!htbp]
\centering
\includegraphics[width=6in]{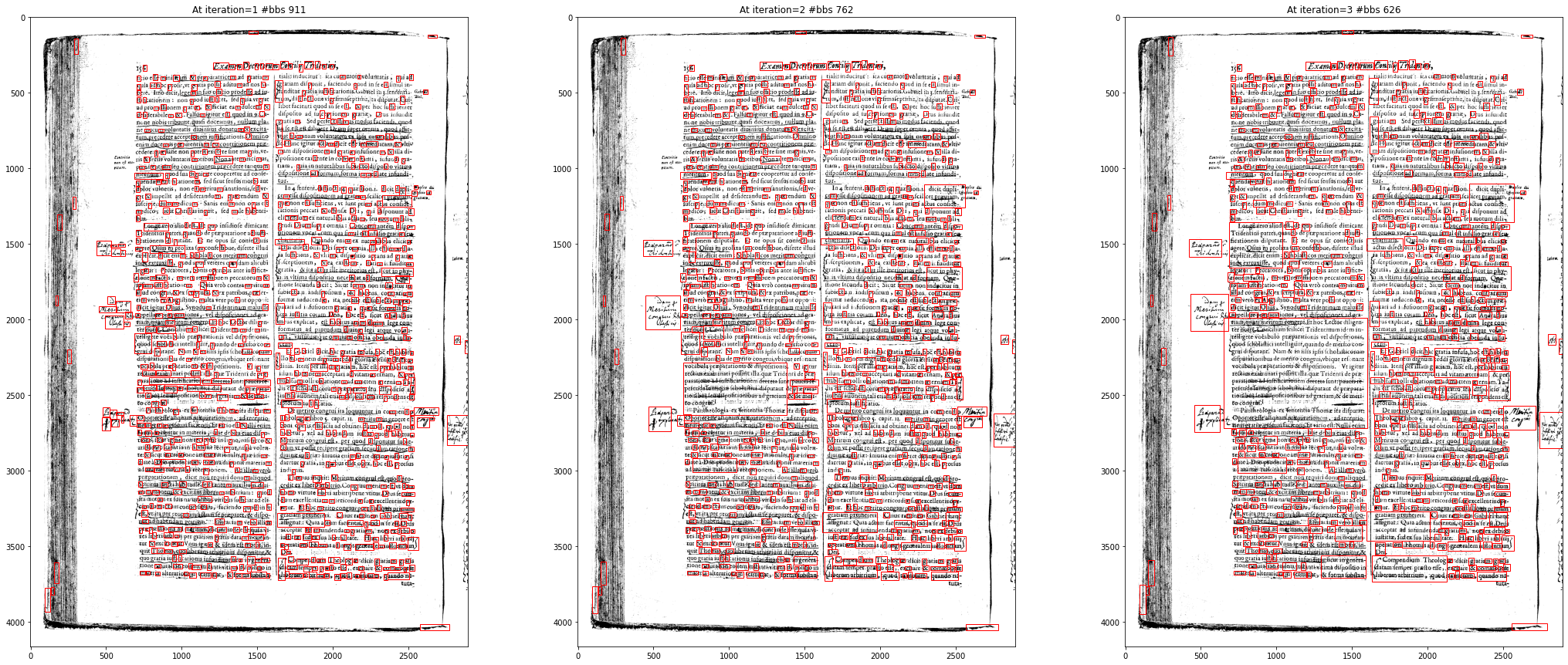}
\caption{Result of MSER with iteration=3, $\Delta$=1.1, i=0.1}
\label{fig:rcnn_mser}
\end{figure*}

\section{Methodology}
The methodology is divided into three parts: localizing the marginalia (using R-CNN and Faster R-CNN); segmenting the found text; and attention-based text recognition pipeline. In the following section, the methods that have been used to accomplish these functionalities will be presented.

\subsection{Data Preprocessing}
As part of the project, data was collected from the early book collections at Uppsala University Library that contain marginalia. The data was prepared and labelled by an expert using an open-source labelling tool \textit{LabelMe} \citep{russell2008labelme}, where for an input image, the marginalia bounding box coordinates were obtained. A total of $513$ labeled images were given, and they were randomly divided into training and test sets in the ratio of 9:1. Data augmentation was performed to supplement the training set, which involves flipping each training image horizontally, adding Gaussian noise, and randomly changing brightness or contrast. With data augmentation, the training set size was increased to $1848$. In addition, the size of each image was re-scaled to 350*500 to reduce the computational cost.

\subsection{Marginalia Detection: R-CNN}
R-CNN uses cropped images, also known as regions of interest (ROI), as input. In localizing marginalia, R-CNN is a classification model in which the input images are classified into two categories: marginalia and non-marginalia, and AlexNet \citep{krizhevsky2012imagenet} is used as the network structure.

The first step in R-CNN experiments is to generate training samples, including ROI of marginalia and non-marginalia. To create ROI of marginalia, the pre-marked bounding box of marginalia is used. We first cut the marginalia parts from the image based on the bounding box coordinates, cropped it into different pieces according the the ratio of length and width, and resized them into 227*227. As for generating ROI of non-marginalia, all the images are processed by Maximally Stable Extremal Regions (MSER) algorithm \citep{matas2004robust}. This algorithm is based on the concept of watershed: binarizing the image between the threshold $[0, 255]$, then the image would go through a process from completely black to completely white. In this process, the area of some connected regions change subtly with the increase of the threshold, and this kind of region is called MSER. 
\begin{equation}
v_{i}=\frac{\left|Q_{i+\Delta}-Q_{i-\Delta}\right|}{\left|Q_{i}\right|}   
\end{equation}
where $Q_i$ represents the area of the i-th connected region; $\Delta$ indicates a small threshold change (water injection); when $v_i$ is less than a given threshold, the region is considered to be MSER. The segmentation of one image is shown in Figure \ref{fig:rcnn_mser}.

After the bounding boxes are obtained, the tiny boxes are removed at first. Then \textit{Intersection over Union}, or \textit{IoU} is calculated for the rest of the bounding boxes, and 4 different boxes with $IoU=0$ are selected as non-marginalia training samples for every image. Applying the same crop and resize procedure as marginalia part to acquire ROI of non-marginalia as the input for model training.

After the preparation of training samples is finished, the next step is to construct the neural network. This R-CNN model uses AlexNet as the network structure, which includes 1 input layer, 5 convolutional layers, 2 fully connected layers and 1 output layer. The basic structure of this network is similar to the original AlexNet \citep{krizhevsky2012imagenet}, only the number of neurons on the fully connected layers and output layer are changed to 500, 20 and 2 according to the number of categories in the actual dataset.

\subsection{Marginalia Detection: Faster R-CNN}
R-CNN has drawbacks, most of all an expensive region proposal computation due to the selective search algorithm. To overcome this run-time issue, we investigated Faster R-CNN \citep{ren2015faster} that combines a region proposal network (RPN) with a Fast R-CNN. The Fast R-CNN runs a fully convolutional network to map an image into a lower resolution spatial feature map. Then, a region of interest (ROI) pooling operator converts each proposed region into a fixed dimensional representation which is the input to a neural network that predicts the object category and the box regression. 

Instead of using the selective search algorithm, Faster R-CNN applies RPN, which is a fully convolutional network that generates region proposals with different scales and aspect ratios. The RPN can be trained and therefore produces better region proposals than the selective search algorithm in the R-CNN. It runs a sliding window over a feature map and determines for different anchor boxes whether there is an object. Further, it predicts deltas to the anchor box to improve its fit. It is a single, unified network that is computationally less expensive than a R-CNN, as the RPN and Fast R-CNN networks share the convolutional computations. %The complete architecture of the Faster R-CNN can be seen in figure \ref{fig:faster_r_cnn}. 

In our implementation, we use the pre-trained ResNet-50 as a network structure for the Faster R-CNN. As we are only interested in one object category, the network predicts $2$ classes, marginalia or non-marginalia, as well as the $4$ box coordinates. We trained the Faster R-CNN for $13$ epochs and a learning rate of $0.001$.

%\begin{figure}[!t]
%    \centering
%    \includegraphics[width=2in]{network.png}
%    \caption{Faster R-CNN network architecture \citep{ren2015faster}. RPN proposes possible object regions.}%
%    \label{fig:faster_r_cnn}%
%\end{figure}

\begin{figure}[!t]
 \begin{minipage}[b]{0.45\textwidth}
    \centering
    \includegraphics[width=3in]{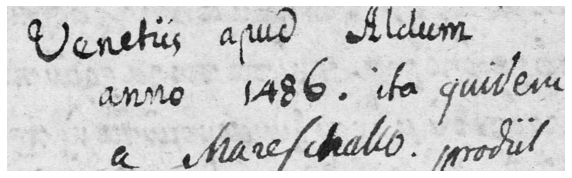}
    \caption{A sample image of marginalia found by the model.}
    \label{fig:1a}
 \end{minipage}
 \begin{minipage}[b]{0.45\textwidth}
    \centering
    \includegraphics[width=3in]{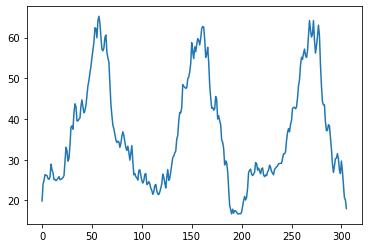}
    \caption{The result of horizontal projection on the image from Figure \ref{fig:1a}}
    \label{fig:1b}
 \end{minipage}
 \end{figure}

  \begin{figure}[!t]
 \begin{minipage}[b]{0.45\textwidth}
    \centering
    \includegraphics[width=3in]{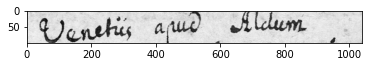}
    \caption{Row 1 obtained by line segmentation}
    \label{fig:2a}
 \end{minipage}
 \begin{minipage}[b]{0.45\textwidth}
    \centering
    \includegraphics[width=3in]{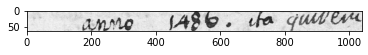}
    \caption{Row 2 obtained by line segmentation}
    \label{fig:2b}
 \end{minipage}
  \begin{minipage}[b]{0.45\textwidth}
    \centering
    \includegraphics[width=3in]{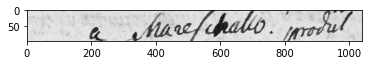}
    \caption{Row 3 obtained by line segmentation}
    \label{fig:2c}
 \end{minipage}
 \end{figure}

  \begin{figure*}[!htbp]
    \centering
    \includegraphics[width=5in]{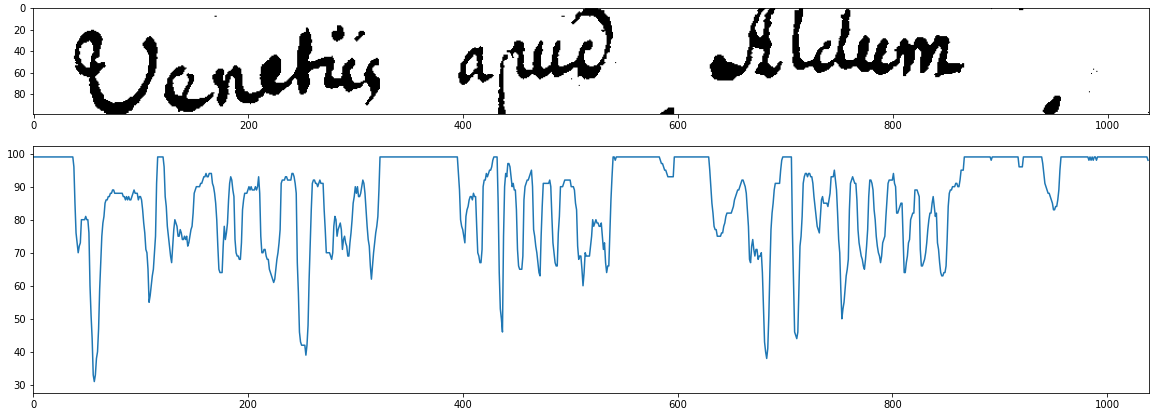}
    \caption{A binarized line and the result of applying vertical projection to it.}
    \label{fig:3}
\end{figure*}

 \subsection{Marginalia Segmentation}
To segment the handwritten marginalia into individual words for further processing, each line of the text must first be identified. To do so, a \textit{sobel} filter is used to emphasize the edges on an image. This is followed by a horizontal projection to the image, which is the sum of pixels on each pixel row. An example image and the result of this projection can be seen in figures \ref{fig:1a} and \ref{fig:1b} respectively.

The peaks that can be seen in the horizontal projection in Figure \ref{fig:1b} shows us where the lines of text are located. This information is used to crop the image to individual lines. A threshold must be set to decide how the rows should be divided. For this implementation the threshold is set to the value in the middle between the highest peak and the lowest value. In figures \ref{fig:2a}, \ref{fig:2b} and \ref{fig:2c}, the resulting rows from the sample image can be seen.

\begin{figure}[!t]
    \centering
    \includegraphics[width=2cm]{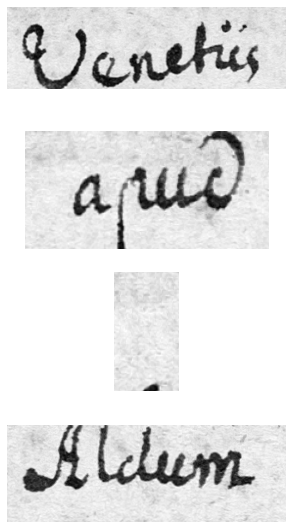}
    \caption{Word segmentation applied to Figure \ref{fig:2a}}
    \label{fig:4}
\end{figure}

Identification of the individual lines is followed by the identification of words. To begin with, the image containing the line is binarized and a vertical projection is applied. This allows observation of spaces in the lines, as can be observed in Figure \ref{fig:3}. In the result from the projection one can see where the spaces between each word are located. However, since a word often has spaces between the letters, it is not suitable to divide on every empty space that is found. To solve this problem, the average length of all spaces is calculated and the line is only split on spaces that are larger than this average. An example of applying this to the line from Figure \ref{fig:2a} can be seen in Figure \ref{fig:4}.

\subsection{Attention-based Recognition}
\begin{figure}[!t]
    \centering
    \includegraphics[width=2.5in]{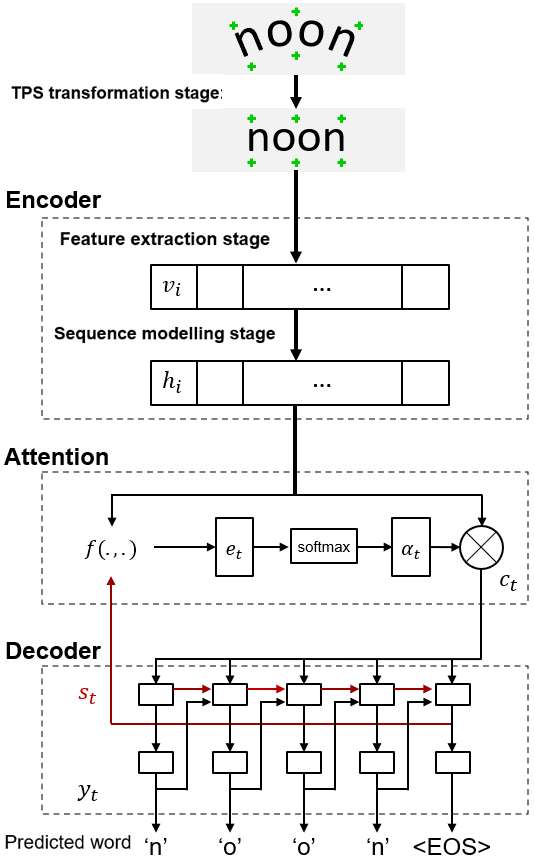}
    \caption{AttentionHTR \citep{kass2022attentionhtr}.}
    \label{fig:attnHTR}%
\end{figure}

After detecting the marginalia and segmenting the words, the goal is to correctly recognise these words. To do so, an end-to-end attention-based sequence-to-sequence model \textit{AttentionHTR} introduced in \citep{kass2022attentionhtr} is used. The model architecture is presented in Figure \ref{fig:attnHTR}, that consists of four stages: thin-plate spline (TPS) transformation, 32 layer ResNet based feature extraction, 2 layer bidirectional LSTM-based sequence modeling, and content-based attention mechanism for prediction. An attention-based decoder is used to improve character sequence predictions. The decoder is a unidirectional LSTM and attention is content-based. The segmented images of words are given as input into the AttentionHTR network, which produces the predicted word along with the confidence scores.

The main advantage of using AttentionHTR model for marginalia recognition is that the general purpose pre-trained model is able to cope with challenging examples due to insufficient annotated data. This is because to handle training data scarcity, AttentionHTR leverages transfer learning from scene images to handwriting images, and uses a  multi-writer dataset (Imgur5K) that contains word examples from 5000 different writers. The architecture is modular and the integration with our pipeline was feasible and computationally inexpensive. 

\begin{figure*}[!t]
    \centering
    \includegraphics[width=6in]{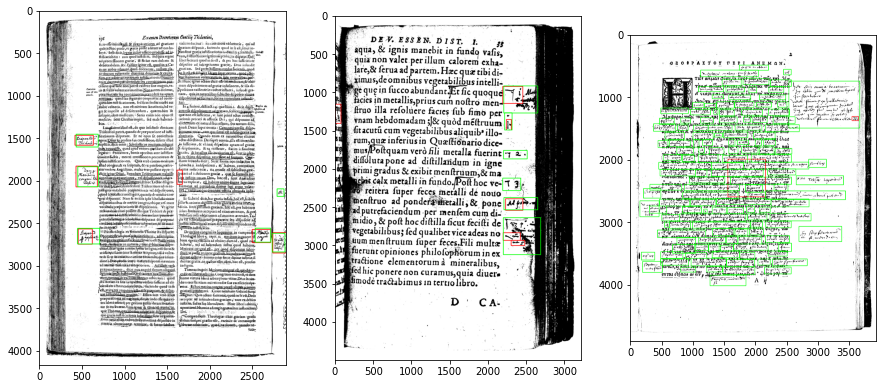}
\caption{Visualization result: Prediction of marginalia on 3 test samples using a R-CNN.}
\label{fig:test_samples}
\end{figure*}

\section{Results}
\subsection{Marginalia Detection}
Table \ref{tab:l1} shows the prediction accuracy and training loss of R-CNN on validation set after each epoch. The accuracy is calculated based on whether the predicted label (marginalia and non-marginalia) matches the pre-marked label.

\begin{table}[!t]
\centering
\begin{tabular}{c|c|c}
\hline
\textbf{Epoch} & Accuracy( \% ) & Loss \\
\hline
\textbf{1} & 85.34 & 0.2120 \\
\textbf{2} & 89.35 & 0.1828 \\
\textbf{3} & 89.04 & 0.1257 \\
\textbf{4} & 89.19 & 0.138 \\
\textbf{5} & 90.74 & 0.2123 \\
\hline
\end{tabular}
\caption{R-CNN: Prediction accuracy and training loss}
\label{tab:l1}
\end{table}

As can be observed in Table 1, the model has been well trained after the third epoch. Although the accuracy on validation looks good, the result of the test samples needs improvement. The general situation is summarized into three categories, as shown in the Figure \ref{fig:test_samples}. In the figure, the green boxes are the originally marked marginalia and the red boxes are the predicted marginalia. The first category: the model is capable to make proper prediction though the predicted boxes don't $100\%$ match the pre-marked boxes. The second category: the model is only able to mark down part of every marginalia bounding box. This is probably due to the input of this model is the small cropped image instead of the full image, the segmentation in the beginning may not cut down the whole marginalia. And the last category (the extreme case), for example, the majority of the image is marginalia. Under such kind of situation, the parameters in segmentation algorithm needs to be changed greatly, otherwise the image can't be segmented well.

% Results Faster R-CNN
\begin{figure}[!t]
    \centering
    \includegraphics[width=3in]{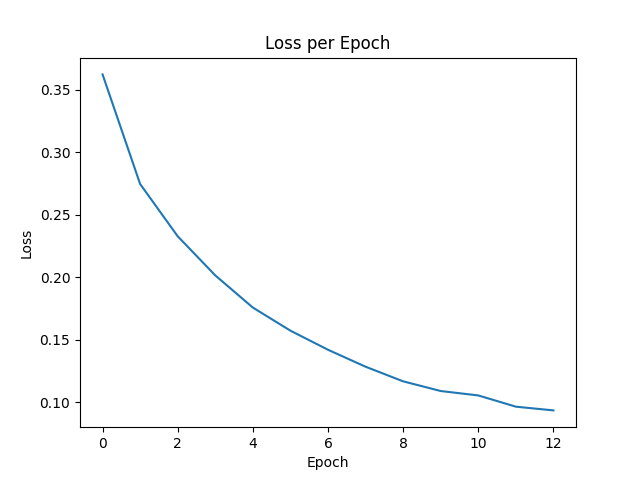}
    \caption{Faster R-CNN training loss.}%
    \label{fig:faster_rcnn_loss}%
\end{figure}

\begin{figure*}[!t]
    \centering
    \includegraphics[width=6.5in]{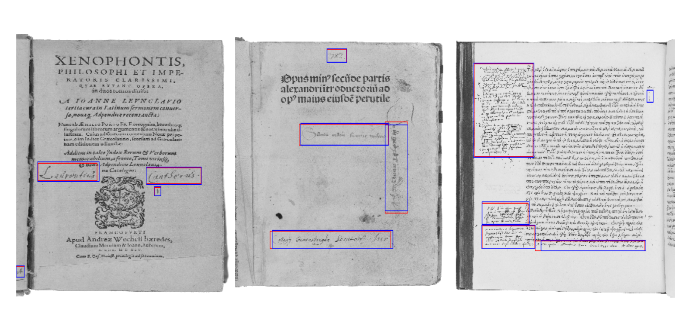}
    \caption{Visualization result: Prediction of marginalia on $3$ test samples using the Faster R-CNN.}%
    \label{fig:faster_rcnn_predictions}%
\end{figure*}

The second set of experiments involve training a Faster R-CNN and applying on the test dataset. The training loss is plotted in Figure \ref{fig:faster_rcnn_loss}. Compared to the R-CNN, the Faster R-CNN performs better and shows a more accurate marginalia detection. An example of predicted marginalia is shown in Figure \ref{fig:faster_rcnn_predictions}, where the labeled marginalia are marked in blue and the predicted marginalia in red. It can be observed that the Faster R-CNN is able to correctly fit the bounding boxes on the marginalia and to label them correctly in most cases. This is the case for different test images, such as images with large marginalia or images containing figures. 

The good performance of the Faster R-CNN is also reflected by a high IoU score of \textbf{0.82}, which indicates that the predicted bounding boxes overlap with the labeled bounding boxes by a large amount. 

%This result is expected, as the Faster R-CNN is the current state-of-the-art network for object detection and an improvement of the R-CNN architecture. 

%Therefore, due to the better performance of the Faster R-CNN over the R-CNN, we will use the Faster R-CNN network for finding marginalia.

\subsection{Marginalia Segmentation}
\begin{figure}[!t]
    \centering
    \includegraphics[width=3in]{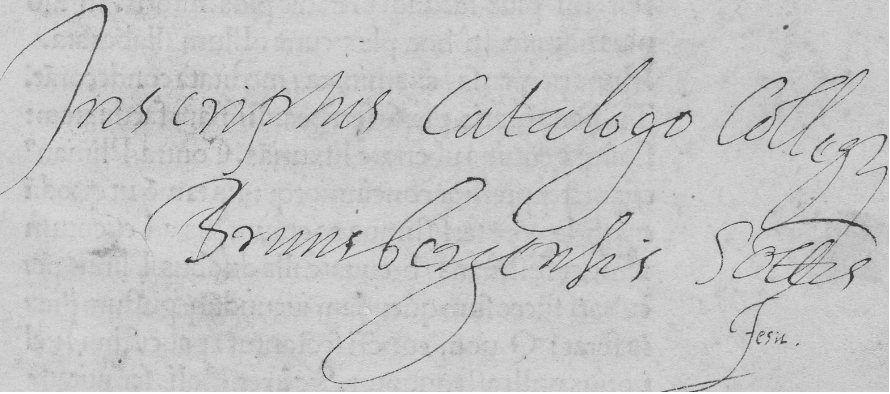}
    \caption{Handwriting with intersecting rows}%
    \label{intersect}%
\end{figure}

Due to the way that the segmentation algorithm is constructed, the results are highly dependant on that the rows do not cross over each other, such as in the example in Figure \ref{fig:1a}. However, this is not always the case since each author has a unique handwriting. One example of marginalia that has this problem can be seen in Figure \ref{intersect}. Since letters from two different lines are located at the same pixel rows the algorithm interprets the entire marginalia to be a single line. 

\begin{figure}[!t]
    \centering
    \includegraphics[width=3in]{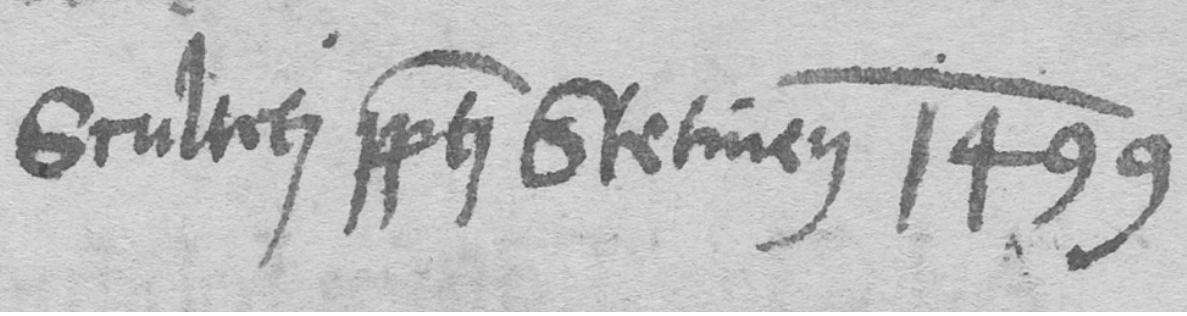}
    \caption{Handwriting with intersecting words}%
    \label{intersect_words}%
\end{figure}

\begin{figure}[!t]
    \centering
    \includegraphics[width=3in]{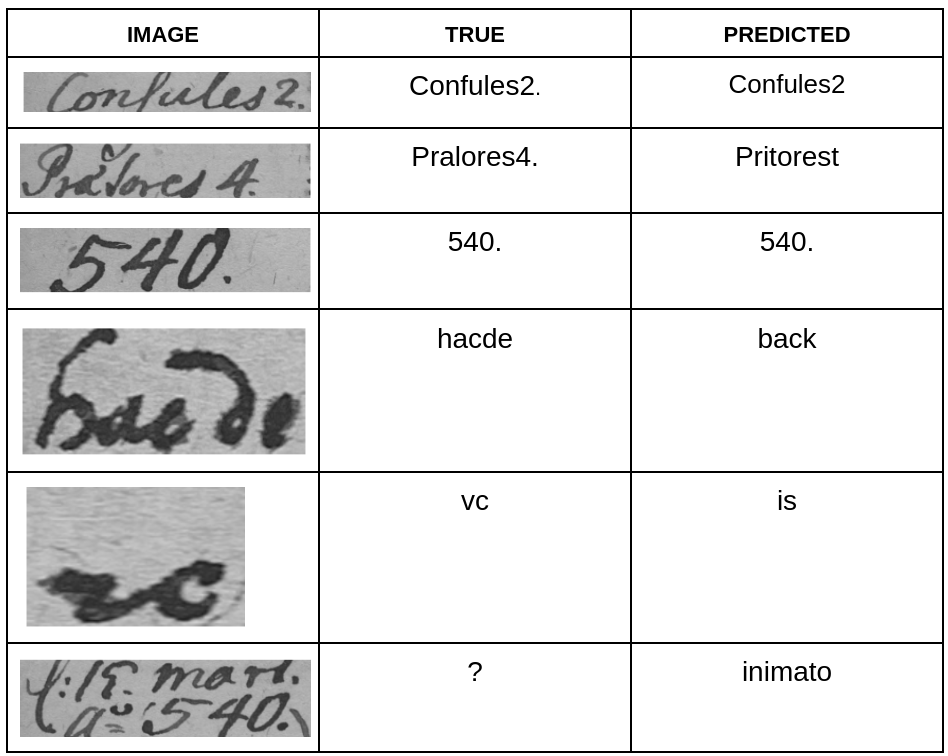}
    \caption{AttentionHTR word recognition results.}%
    \label{fig:AttentionHTR}%
\end{figure}

Similarly, it is also important that separate words on the same line do not intersect. This problem does not seem to be as frequently occurring but it is still something that could be mentioned. An example is presented in Figure \ref{intersect_words}, where there is a line that is drawn above the middle word and the number at the right-hand side. Due to this line, the algorithm as it is constructed at the moment, is not able to separate the two words. A different method would be required to solve such problem. 

\subsection{Text Recognition using AttentionHTR}
The segmented words were given as input into the AttentionHTR network.
A selection of these results can be seen in Figure \ref{fig:AttentionHTR}, where the first column contains the input image, the second column contains the true word, and the third column presents the predicted word by AttentionHTR. 
As one can see, AttentionHTR labels the words and even numbers very reasonably. However, some of the words provided as input are not even readable for the human eye. As expected, AttentionHTR is not able to correctly label those input images. Further, it struggles if the input image is not perfectly segmented. For example, if the input image contains multiple words in different lines, the predicted text is inaccurate. Overall, AttentionHTR performs very well if the input image is clear, well segmented, and also readable for the human eye.

\section{Discussion}
%\subsection{Marginalia Detection}
% RCNN
In general, the R-CNN model works well if the input images have been well-cut. But the segmentation is not always perfect because the layout of each image is different. For instance, some have multiple marginalia outside of the printed text, some have a few marginalia inside the printed part, some have graphics besides handwritten and printed text, etc. The difference between each image requires adjustment of segmentation algorithm parameters for each image to make a good segmentation. Therefore, only with a better way to do the segmentation for an input image, can this model perform better. 

In contrast to this, Faster R-CNN does not require the pre-segmentation operation, so it has a good chance to address the problem of not marking all potential marginalia before the images are input to the model. Instead, the regional proposal network of the Faster R-CNN does not only speed up computation, but also seems to propose more relevant sections of the image. Combined with the Fast R-CNN, the coordinate predictions for the marginalia are much more accurate than the ones of the R-CNN. The accuracy, together with the more efficient computation, lets the Faster R-CNN outperform the R-CNN. This is also reflected by a high IoU score (\textbf{0.82}) of the Faster R-CNN. 

%\subsection{Marginalia Segmentation}
With regard to the marginalia segmentation, it was observed that the word segmentation algorithm struggles with lines and words that intersect with each other. These problems are very hard to solve with the current implementation, and requires more sophisticated algorithms. It would be interesting to try a R-CNN approach similar to the method for localizing the marginalia. However, this would require a good amount of time dedicated to creating training data for the algorithm by manually labeling coordinates of the words.

While the Faster R-CNN was found to be well performing for our use-case, there also exists other object detection methods that were not explored in this study, such as Mask R-CNN \citep{he2017mask} and YOLO \citep{redmon2016you}. However, the dataset is prepared in a format easily adaptable for different architectures, and other architectures such as Mask R-CNN and YOLO can be investigated as a potential future work. For text recognition, Transformer based methods such as TrOCR \citep{li2023trocr} also exists, but this study focused on AttentionHTR due to a powerful in-house general purpose HTR multi-writer model, developed as part of our previous work \citep{kass2022attentionhtr}. The experimental evaluation can be further extended with a comparison with other text recognition methods as future work. 

%\subsection{Text Recognition}
The results show that the text recognition accuracy with AttentionHTR heavily depends on the quality of the segmented input image. If the image is clear and well segmented, i.e., only one word is visible, AttentionHTR provides accurate results, without the need for training from scratch. However, some output images of the marginalia segmentation algorithm were multiple lines long, or the quality of the given image was just not precise enough. In that latter case, even a human cannot correctly classify the text. Overall, AttentionHTR is highly accurate when the input was clear and well segmented. In future work, words in the dataset can be labeled, which would allow for training of the AttentionHTR network on the given data. Also, the present study will be extended further to include annotated marginalia such as Melville \citep{norberg2023technical} and Mill \citep{pionke2020handwritten}. 

%Reviewer 2: why a language model was not included. why existing annotated marginalia (e.g., Melville, Mill etc.) would not be useful for this study, since the current training data didn't include annotated marginalia content at all.

\subsection{Conclusion}
This work presented an end-to-end framework for automatic detection and recognition of handwritten marginalia. The experimental results on the data from Uppsala University Library's early book collections demonstrate the effectiveness of the proposed method, where Faster R-CNN was found to perform better than R-CNN for marginalia localization, and AttentionHTR contributed towards the recognition performance. Under the experimental settings, the proposed HTR pipeline produced encouraging results for both marginalia detection and recognition. However, since the training data is limited and expert knowledge is needed for annotating the marginalia texts, the future work involves collaboration with the librarians and professionals to prepare training data for recognition of historical marginalia texts, written in both Swedish and English. Furthermore, language modeling, a different regularization method, and generative AI models for handwriting synthesis will be explored. The source code and pre-trained models are made available for advancing the research further at \href{https://github.com/adamaxelsson/Project-Marginalia}{GitHub}.

\section*{Limitations}
Some of the limitations of this work include: 

\begin{itemize}
    \item The training data used in this work contains labeled bounding box coordinates for marginalia, but not the annotations representing what the marginalia reads. Unavailability of annotated marginalia is one of the main limitations of this work, making it challenging for the model to generalise on unseen data. %Moving forward, the right quality and quantity of training data is the key towards better performance. 
    \item It is not straightforward to annotate the marginalia text and is a time-consuming process. It requires expert knowledge, where an expert should be able to read a variety of handwriting, and is multi-lingual as one cannot know beforehand about the language used for marginalia (it can vary with different readers). 
    \item To enhance the recognition performance, the authors have been investigating the integration of a language model (such as Skip-gram) in the pipeline. However, it was found to be more suitable as a post-processing step due to the architectural limitations with CNN-based feature extraction and sequential modeling. Also using a language model such as BERT at post-processing can enhance the accuracy, but it might also increase the computational cost, and the authors will investigate this further as future work. 
    \item A limitation of an end-to-end detection and recognition pipeline is that it is challenging to tailor or tweak the models to handle cases with special scripts (e.g. Gothic, curlicue), blurry text, strike-through words, poor handwriting, etc. Therefore, we aim to have a general-purpose model that is good enough for a variety of data.
\end{itemize}

\section*{Ethics Statement}
The research conducted in this work respects the rights and welfare of the society, general public and other stakeholders such as librarians, historians and research scholars. The languages studied herein includes Swedish and English, and the methods thus developed can be applied to other languages. The source code and pre-trained models are made public to the research community to benefit the research, future re-use, and for the dissemination of knowledge to the public. The research does not pose any risk or harm to anyone, and is conducted with honesty and integrity.

%Other potential future work includes exploring other network architectures for marginalia predictions. While the Faster R-CNN led to highly accurate results, there are also other object detection methods that are to be explored, such as Mask R-CNN \citep{he2017mask} and YOLO \citep{redmon2016you}. An accurate segmentation of words within the marginalia seems to be key for a correct text recognition with AttentionHTR, and more sophisticated approach towards segmentation is desired.

%Finally, to increase the performance of the text recognition, one can fine-tune AttentionHTR by training in on additional data. This would be possible when the words on the images were labeled. Even though this is a tedious process, the labeling would lead to a more accurate text recognition. In collaboration with the librarians and professionals, our plan is to prepare training data for recognition of historical marginalia texts, written in both Swedish and English. 

\section*{Acknowledgment}
This research was partially supported by \emph{Kjell och M{\"a}rta Beijer Foundation} and \emph{Uppsala-Durham Strategic Development Fund}: "Marginalia and Machine Learning: a Study of Durham University and Uppsala University Marginalia Collections". The computations were enabled by resources provided by the National Academic Infrastructure for Supercomputing in Sweden (NAISS) partially funded by the Swedish Research Council through grant agreement no. 2022-06725. The authors would like to thank Raphaela Heil and Peter Heslin for valuable suggestions and feedback.

\bibliography{anthology,custom}
\bibliographystyle{acl_natbib}

%\appendix

%\section{Example Appendix}
%\label{sec:appendix}

%This is a section in the appendix.

\end{document}